\documentclass[11pt,letterpaper]{article}

\usepackage[T1]{fontenc}
\usepackage[utf8]{inputenc}
\usepackage{lmodern}   
\usepackage{microtype} 

\usepackage[margin=1in]{geometry}
\usepackage{setspace}

\usepackage{amsmath}
\usepackage{amssymb}

\usepackage{booktabs}
\usepackage{graphicx}
\usepackage{caption}
\usepackage{float}

\usepackage{tikz}
\usetikzlibrary{arrows.meta,positioning}

\usepackage{newfloat}
\usepackage{listings}
\DeclareCaptionStyle{ruled}{labelfont=normalfont,labelsep=colon,strut=off}
\DeclareFloatingEnvironment[fileext=lst,name=Listing,placement=tb]{listing}
\usepackage[numbers,sort&compress]{natbib}
\usepackage[hyphens]{url}
\usepackage[colorlinks=true,allcolors=blue!55!black,breaklinks=true]{hyperref}

\title{\bfseries LLM-Generated Feature Pools \\
for Time Series Anomaly Detection}

\author{%
  Youssef {Attia El Hili}\textsuperscript{{\normalfont 1}}\textsuperscript{{\normalfont 2}},\;
  Malik Tiomoko\textsuperscript{{\normalfont 1}},\;
  Corinne Ancourt\textsuperscript{{\normalfont 2}}\\[4pt]
  \normalsize
  \textsuperscript{1} Huawei Noah's Ark Lab, Paris, France\\
  \textsuperscript{2} Centre de Recherche en Informatique, Mines Paris, PSL University\\
  \texttt{youssef.attiaeh@gmail.com}
}
\date{}

\begin{document}

\maketitle

\begin{abstract}
We study how far a simple statistical pipeline can go on univariate time series anomaly detection under a strict selection protocol. The method extracts a small pool of statistics over sliding windows, scores each window with a transductive robust (MAD) model, and selects a feature subset per domain on a held-out tuning split. On TSB-AD-U it reaches $0.529$ per-series VUS-PR, above the best neural ($0.45$) and statistical ($0.44$) entries on the public leaderboard and within $0.06$ of the strongest pretrained foundation model, several of which use more supervision than ours. Ablations locate the cause: across three selection strategies and a hindsight oracle the score moves by $0.031$, and across the aggregation grid by $0.096$, while changing the candidate pool moves it by $0.226$. The candidate pool sets the ceiling; the search over it is second-order. We therefore generate a pool per domain by prompting a multimodal LLM with in-context example windows from that domain. The generated pools match the hand-crafted one under matched selection, and the two cover different domains: selecting over their union improves on the generated pool in all twelve generator--seed pairs and lifts the pipeline to $0.588$, matching the performance of the best entry on the leaderboard.
\end{abstract}

\section{Introduction}

Time series anomaly detection spans a broad range of methods, from simple to complex. At the simple end, distance, density, and statistical detectors score points or windows against a model of normal behaviour built from a few summary features. At the complex end, reconstruction and forecasting models, deep neural detectors, and pretrained foundation models learn that model of normality from data. The usual expectation is that the complex end should dominate, since it can capture richer structure.

Under a common protocol with comparable tuning, however, classical detectors remain competitive with substantially more elaborate ones~\cite{liu2024elephant, paparrizos2022tsb}. If elaborating the detector is not what determines performance, it is worth asking what does.

\begin{figure}[h]
\centering
\includegraphics[width=0.85\columnwidth]{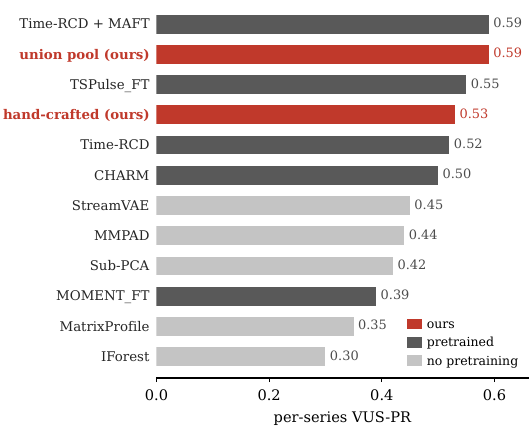}
\caption{Per-series VUS-PR on TSB-AD-U. Our pipeline scores each series with a transductive median/MAD model over a small pool of window statistics. With hand-crafted features it already clears the best neural (StreamVAE) and statistical (MMPAD) entries; with a pool discovered per domain by a language model and unioned with the hand-crafted one, it reaches the level of the best pretrained entry on the leaderboard. Red marks our two variants, dark grey the pretrained foundation models, and light grey the methods without pretraining. Leaderboard values are as published.}
\label{fig:hook}
\end{figure}

For a feature-based detector, much of the answer is fixed before the detector runs. Such a method can only react to what its features encode: if no feature in its pool captures the signature of an anomaly, that anomaly is invisible to the detector, and no scoring rule or search over feature subsets can recover it. The difficulty is that the signature is not one thing. What makes a window anomalous differs in kind across domains, from shifts in level to changes in variance regime, distortions of shape or periodicity, and violations of context that no marginal statistic exposes, so a fixed pool is a bet, placed before the data is seen, on which of these will matter. Any individual blind spot is easy to patch once it has been named, but naming it requires having already studied the domain, and the patch does not transfer to the next one. Nor does the scalable version of patching work: a large generic library is that response taken to its limit, and catch22~\citep{catch22}, curated across thousands of candidate statistics, is the weakest pool we run on this benchmark (Section~\ref{sec:knobs}). A feature-based pipeline is therefore limited less by its detector or its selection procedure than by the vocabulary it is handed, and the way to improve that vocabulary is to condition it on the problem rather than to enlarge it.

This paper targets the vocabulary directly. The result (Figure~\ref{fig:hook}) is a transductive, parameter-free detector whose hand-crafted form clears every statistical and neural pipeline on the TSB-AD-U leaderboard and that can be combined with features generated by a multimodal LLM to match the strongest pretrained foundation model.

That vocabulary has usually been static. Feature-based detection relies on fixed, expert-curated libraries of statistics, from broad collections such as catch22~\cite{catch22} and tsfresh~\cite{tsfresh} to smaller hand-tuned pools, on the assumption that a rich enough generic set transfers to any task. We adopt a deliberately minimal pipeline in this tradition: sliding-window statistics, a transductive robust median/MAD detector fit on each series alone, and a feature subset chosen per domain on a held-out tuning split, with the evaluation split never seen during selection. It has no learned parameters beyond per-series statistics, yet under this strict protocol it reaches $0.529$ per-series VUS-PR on TSB-AD-U, well above the best neural ($0.45$) and statistical ($0.44$) entries on the public leaderboard and within a few points of pretrained foundation models, which are the only method family that beats it.

Its performance is bounded by the pool, not by the selection or aggregation one would normally tune. Across three selection strategies the score moves by three points, and a per-domain oracle chosen with hindsight is barely better; sweeping the score-aggregation grid moves it by ten. Changing the candidate pool moves it by twenty-three (Figure~\ref{fig:knobs}). The obvious response, a larger curated library, does not help either: catch22, more than twice the size of our pool, scores lower, because a static vocabulary cannot encode what it was not designed to capture. Enlarging the static set is the wrong response.

The productive axis is orthogonal: from a fixed vocabulary to feature discovery conditioned on the problem. We prompt a multimodal language model with example windows from a domain and ask it to propose problem-specific feature functions, then run the same selection protocol on the generated pool. Language models are natural tools for adaptive feature discovery because they can synthesize new representation hypotheses from only a few examples of the target domain. We find that the stronger generators are competitive on their own, with no manual feature engineering. A language model alone is not uniformly reliable, so the real gain is complementarity. Expert priors and discovered features fail on different domains, so selecting over their union beats either alone across all three generators and lifts the statistical pipeline to $0.588$, closing the gap with the best entry on the leaderboard.

Our contributions are as follows. We show that a minimal transductive statistical pipeline, under a strict per-domain selection protocol, outperforms the best neural and statistical entries on TSB-AD-U and is competitive with pretrained foundation models. We show, through matched ablations over feature selection and over score aggregation, that the candidate feature pool rather than the selection or the aggregation is the binding constraint, and that enlarging a static library (catch22) does not lift it. We give a procedure that discovers a per-domain pool automatically with a language model, competitive on its own for the stronger generators and complementary to expert features, so that their union reaches the level of the strongest pretrained model on the leaderboard. We frame this as a shift from enlarging static feature libraries to adaptive, task-conditioned feature discovery, and find that feature generation, more than selection, is what limits these pipelines.

\section{Related Work}

\paragraph{Time series anomaly detection and benchmarks.}
The field spans forecasting-based, reconstruction-based, distance-based, and density-based detectors~\cite{paparrizos2022tsb}. A recurring lesson from recent benchmarking efforts is that evaluation protocol and tuning matter as much as method family, and that classical detectors are hard to beat once everyone is tuned consistently~\cite{liu2024elephant, paparrizos2022tsb}. We adopt TSB-AD-U~\cite{liu2024elephant} and its VUS-PR evaluation~\cite{paparrizos2022volume} so that our comparison is on the same footing as published baselines. Pretrained foundation models and deep detectors have advanced quickly and are among the strongest entries on this benchmark~\cite{ansari2024chronos, goswami2024moment, tspulse}.

\paragraph{Static feature libraries.}
A long line of work distills time series analysis into fixed libraries of statistics, from broad collections such as catch22 and tsfresh to smaller curated pools~\cite{catch22, tsfresh}. These libraries embody a static-vocabulary paradigm: they are designed once, independent of any domain or detector, on the bet that a rich enough set of generic features transfers to any task. This is powerful but has a structural limit, since a feature not in the library cannot be recovered by selecting over it. Our hand-crafted pool is a compact instance of the same paradigm, and we use catch22 as its strongest representative when we test whether enlarging the static vocabulary closes the gap.

\paragraph{Language models as feature generators.}
Language models have been applied in context to both ends of the machine learning pipeline, proposing new features for a downstream learner~\cite{CAAFE} and selecting models or hyperparameters from a fixed space~\cite{xu2026treestructuredsynergylargelanguage}. They have also been used to write and refine code against an objective~\cite{FunSearch2023}, and to operate directly on time series~\cite{gruver2023large, jin2023time}. We use a language model in a narrow but different role from a static library: rather than selecting from a fixed vocabulary, it proposes problem-specific representation hypotheses, small feature functions conditioned on examples from the target domain, which are then compiled and scored by a fixed statistical detector. Of the two roles above we use only the first. The model never sees the evaluation split and never scores anomalies itself.

\section{Method}

\subsection{Pipeline overview}

\begin{figure}[t]
\centering
\begin{tikzpicture}[
    node distance=3.2mm,
    box/.style={draw, rounded corners, align=center, inner sep=3pt,
                font=\footnotesize, minimum height=7.5mm, text width=64mm},
    arr/.style={-{Latex[length=2mm]}, thick}
]
\node[box] (a) {Series $\rightarrow$ sliding windows ($w=$ dominant period, stride $w/2$)};
\node[box, below=of a] (b) {Per-window features $X \in \mathbb{R}^{n_w \times d}$};
\node[box, below=of b] (c) {Transductive median/MAD fit, robust $|Z|$ per feature};
\node[box, below=of c] (d) {Feature-agg: $|Z| \rightarrow$ one score per window (mean)};
\node[box, below=of d] (e) {Point-agg: overlapping windows $\rightarrow$ per-point score (mean)};
\node[box, below=of e, fill=black!5] (f) {VUS-PR under the TSB-AD protocol};
\draw[arr] (a) -- (b);
\draw[arr] (b) -- (c);
\draw[arr] (c) -- (d);
\draw[arr] (d) -- (e);
\draw[arr] (e) -- (f);
\end{tikzpicture}
\caption{The pipeline. Steps 1--5 are transductive and label-free; labels enter only in the per-domain feature selection that fixes which columns of $X$ are used (Section~\ref{sec:selection}).}
\label{fig:pipeline}
\end{figure}
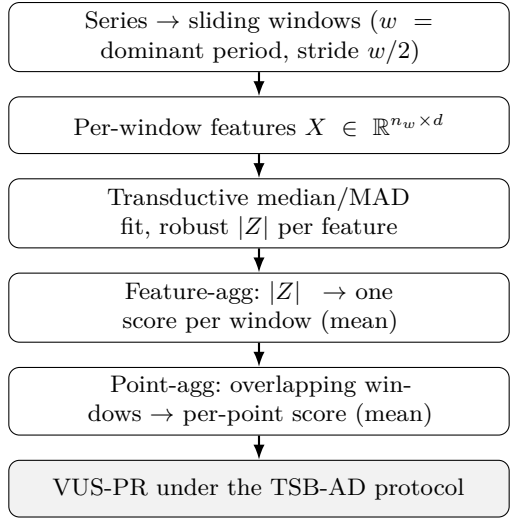

The pipeline maps a single univariate series to a per-point anomaly score in five steps (Figure~\ref{fig:pipeline}). It is fully transductive: every quantity is computed from the series being scored.

\begin{enumerate}
\item \textbf{Windowing.} We slide a window of length $w$ with stride $s$ over the series. By default $w$ is the series' own dominant period, estimated without labels, and $s = \lfloor w/2 \rfloor$.
\item \textbf{Feature extraction.} For each window we compute a fixed vector of statistical features, giving a matrix $X \in \mathbb{R}^{n_w \times d}$ of $n_w$ windows and $d$ features.
\item \textbf{Robust scoring.} We fit a median/MAD model per feature column on $X$ and compute a robust z-score $|Z|$ for every entry.
\item \textbf{Feature aggregation.} We reduce $|Z|$ across features to one score per window (default: mean).
\item \textbf{Point aggregation.} We map overlapping window scores back to per-point scores by averaging the windows that cover each point.
\end{enumerate}

\subsection{Robust detector}

Let $X_{:,j}$ be the $j$-th feature column over the $n_w$ windows of a series. We fit
\[
m_j = \mathrm{median}(X_{:,j}), \quad
\mathrm{mad}_j = \mathrm{median}\big(|X_{:,j} - m_j|\big),
\]
replacing any $\mathrm{mad}_j = 0$ with $1$ to avoid division by zero, and score each entry with $|Z_{ij}| = |(X_{ij} - m_j)/\mathrm{mad}_j|$. The per-window score is the mean of $|Z_{ij}|$ across the selected features. The median/MAD estimator is standard in robust statistics because it tolerates a substantial fraction of contaminated points~\cite{MAD}, which suits a setting where anomalies are present but unlabelled at scoring time. The detector has no learned weights: fit and score both reduce to per-column medians on the target series.

\subsection{Handcrafted feature pool}

The hand-crafted pool contains ten features chosen to be cheap and interpretable: standard deviation, interquartile range, skewness, kurtosis, mean absolute change, mean absolute second derivative, zero-crossing count, lag-1 and lag-2 autocorrelation, and a histogram entropy. Each is a pure function of one window returning a finite scalar. This pool is intentionally small; a central question of the paper is whether a comparable or better pool can be produced automatically.

\subsection{Feature selection}
\label{sec:selection}

Selection is the only place labels are used, and it happens strictly on the tuning split. For a domain with tuning series $\{(x_k, y_k)\}$, the objective of a candidate subset $S$ is the mean VUS-PR obtained by scoring each tuning series with the detector restricted to $S$. During selection we use a fast AUC-PR proxy for this objective to keep the search cheap; the reported evaluation scores always use full VUS-PR.

We consider three selection strategies: greedy forward selection, top-$k$ by single-feature relevance, and mRMR-style redundancy-penalised selection. Each strategy is run on the tuning split. Greedy forward selection evaluates $O(d^2)$ candidate subsets, which is affordable for a ten-feature pool but not for the larger ones, so for pools of twenty features or more, namely catch22 and the union pool, we run only top-$k$ and mRMR. Section~\ref{sec:knobs} shows that this restriction does not drive any of the comparisons.

\subsection{Adaptive feature discovery with a language model}
\label{sec:generation}

\begin{figure}[t]
\centering
\begin{tikzpicture}[
    node distance=2.6mm,
    box/.style={draw, rounded corners, align=center, inner sep=3pt,
                font=\footnotesize, minimum height=6.8mm, text width=63mm},
    hl/.style={draw, rounded corners, align=center, inner sep=3pt, fill=black!5,
                font=\footnotesize, minimum height=6.8mm, text width=63mm},
    arr/.style={-{Latex[length=2mm]}, thick}
]
\node[box] (a) {Tuning windows per domain, split into anomalous vs.\ normal};
\node[box, below=of a] (b) {Rendered as plots on a shared axis};
\node[hl, below=of b] (c) {Multimodal LLM $+$ narrow instruction};
\node[box, below=of c] (d) {Ten fenced Python feature functions};
\node[box, below=of d] (e) {Compile, run on probe windows, drop invalid};
\node[box, below=of e] (f) {Generated pool $\rightarrow$ \emph{same} selection protocol};
\draw[arr] (a) -- (b); \draw[arr] (b) -- (c); \draw[arr] (c) -- (d);
\draw[arr] (d) -- (e); \draw[arr] (e) -- (f);
\end{tikzpicture}
\caption{Automatic pool generation (Section~\ref{sec:generation}). The model sees only rendered tuning windows and a fixed instruction; it never sees the evaluation split, the detector, or any score. Everything downstream of the pool is identical to the hand-crafted pipeline, so the candidate pool is the only variable that changes.}
\label{fig:gen}
\end{figure}
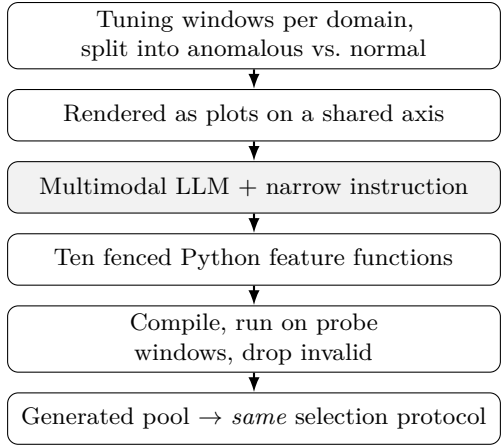

Instead of selecting from a fixed vocabulary, we let a language model propose representation hypotheses conditioned on the problem. To build a pool for a domain, we first sample a small set of example windows from that domain's tuning series, split into windows that contain a labelled anomaly and windows that do not (Figure~\ref{fig:gen}). Each window is rendered as a line plot on a shared y-axis, so that level and magnitude differences between windows are visible, and the plots are passed to a multimodal language model together with a short instruction. We show the model only a handful of examples per class; it never sees the evaluation split, the detector, or any score. To limit the risk that a pretrained model recalls domain-specific features it may have seen during pretraining, we never name the domain or dataset in the prompt and pass only the window plots, so the model must reason from the visible patterns rather than from prior knowledge of the benchmark.

The instruction is deliberately narrow. It states the task (separate anomalous from normal windows with a robust median/MAD detector), points the model at the failure modes it can read off the plots, and constrains the output to compilable, self-contained code. The operative part of the prompt is:
\begin{quote}\footnotesize
Propose exactly ten Python feature functions that would help a robust (median/MAD) detector separate anomalous windows from normal ones in this domain. Study the plots and target the failure modes you see (level shifts, spikes, variance changes, shape or periodicity changes). Each function takes a 1-D window \texttt{x}, uses only \texttt{np}, returns a finite float, and is robust to windows as short as two samples.
\end{quote}

The returned code is parsed from a single fenced block, executed in a restricted namespace with numpy exposed as \texttt{np}, and each function is validated on a set of probe windows; functions that raise, that require imports, or that always return non-finite values are discarded. The surviving functions form the generated pool. Listing~\ref{lst:humanactivity} shows a representative output. The example windows are drawn with a fixed random seed, and different seeds yield slightly different pools, which we exploit to report seed variability.

\begin{listing}[t]
\begin{lstlisting}[language=Python]
def feat_mean_level(x):
    return float(np.mean(x))
def feat_median_level(x):
    return float(np.median(x))
def feat_min_level(x):
    return float(np.min(x))
def feat_q25_level(x):
    return float(np.percentile(x, 25))
def feat_ratio_above_minus_hundred(x):
    return float(np.mean(x > -100))
def feat_median_absolute_deviation(x):
    med = np.median(x)
    return float(np.median(np.abs(x - med)))
\end{lstlisting}
\caption{Six of the ten functions generated for the \emph{human activity} domain (\texttt{gemini-3.5-flash}). The model read level shifts off the plots and proposed a battery of location statistics. The hand-crafted pool contains no such feature, which is why it cannot see these anomalies (Table~\ref{tab:perdomain}).}
\label{lst:humanactivity}
\end{listing}

We compare several feature sources under this fixed procedure. The \texttt{handcrafted} source uses our ten-feature pool. The \texttt{catch22} source uses the 22 features of the catch22 library~\cite{catch22}, a strong static representative curated across thousands of candidates, as an off-the-shelf pool. The \texttt{llm} source uses the generated pool alone. The \texttt{union} source selects over the concatenation of the generated and hand-crafted pools, roughly twenty candidate features. Everything downstream, the detector, the windowing, the selection protocol, and the evaluation, is held fixed across sources; only the candidate pool changes.

\section{Experimental Setup}
\label{sec:setup}

\paragraph{Data.}
We evaluate on TSB-AD-U, the univariate track of the TSB-AD benchmark~\cite{liu2024elephant}, which groups series into nine domains: environment, facility, finance, human activity, medical, sensor, synthetic, traffic, and web services. The benchmark provides a fixed split of 48 tuning series and 350 evaluation series. Throughout, feature selection and any window-length tuning are performed on the tuning split of each domain in isolation, and the 350 evaluation series are scored exactly once, after all design choices are frozen. This protocol is the same one applied to the published baselines, so the comparison is like for like.

\paragraph{Evaluation metric.}
We report VUS-PR~\cite{paparrizos2022volume}, the volume under the precision-recall surface, computed under the TSB-AD protocol in which the evaluation window is set to each series' estimated period. VUS-PR is threshold-free and tolerant to small localisation errors, which makes it the standard choice for this benchmark. We aggregate into a single per-series mean over all 350 evaluation series, weighting each domain by its number of series, which is the aggregate reported in the benchmark's own tables and on its leaderboard.

\paragraph{Baselines.}
We position the pipeline against the public TSB-AD-U leaderboard, taking the best entry in each method family: MOMENT\_FT~\cite{goswami2024moment}, Time-RCD\,+\,MAFT~\cite{timercd}, CHARM~\cite{CHARM}, and TSPulse\_FT~\cite{tspulse} among pretrained foundation models, StreamVAE~\cite{streamvae} among neural detectors, and MMPAD~\cite{mmpad} among statistical ones, together with MatrixProfile~\cite{matrixprofile}, Sub\_PCA~\cite{liu2024elephant}, and IForest~\cite{iforest} for context. These values are reported as published.

Several baselines use more supervision than our pipeline: the foundation models are pretrained on external corpora and finetuned on the tuning set. Our detector has no trained parameters and consults tuning labels only in the per-domain feature-selection step and generation when an LLM is used.

\paragraph{Generators.}
{\sloppy
Pools are generated by three multimodal models, \texttt{gemini-3.5-flash}~\cite{gemini35flash}, \texttt{gemini-3.1-flash-lite}~\cite{gemini31flashlite}, and \texttt{claude-sonnet-5}~\cite{ClaudeSonnet52026}, using the procedure of Section~\ref{sec:generation}, each accessed through its provider's API. For each model we generate an independent pool under four random seeds, which vary the sampled example windows, and report the mean over seeds together with the population standard deviation across seeds. All generated-pool runs use the period-based window and mean/mean score aggregation.\par}

\paragraph{Score aggregation and comparability.}
The pipeline reduces per-feature robust scores $|Z|$ to a per-window score (feature aggregation) and then combines overlapping windows into per-point scores (point aggregation). Sweeping the full pool without selection over this grid, mean/mean ranks first, and we use it for every result in the paper, so all comparisons are on matched aggregation.

\section{Results}

We organise the results as a progression. First, the hand-crafted pipeline clears every method on the leaderboard that, like it, is trained on nothing outside the series being scored, and trails only pretrained foundation models (Table~\ref{tab:main}). Second, neither the selection strategy nor the score aggregation explains the remaining gap: both move the score far less than the candidate pool does (Figure~\ref{fig:knobs}), and a larger static library, catch22, moves it in the wrong direction (Table~\ref{tab:pools}). Third, adaptively discovered pools are competitive on their own, and selecting over the union of expert and discovered features beats either and closes the gap to the best pretrained entry (Table~\ref{tab:generated}). Fourth, a per-domain analysis shows why: expert and discovered pools fail on different domains, so generation, more than selection, is what limits the pipeline (Table~\ref{tab:perdomain}).

\subsection{Comparison with tuned baselines}

Table~\ref{tab:main} places the hand-crafted pipeline against the TSB-AD-U leaderboard, organised by method family. The pattern is clean at the family level. Against detectors that are, like ours, trained on nothing outside the series being scored, the pipeline is ahead by a wide margin: $0.529$ against $0.45$ for the best neural entry (StreamVAE) and $0.44$ for the best statistical one (MMPAD), a relative gain of roughly a fifth over each. Against pretrained foundation models it is competitive: it sits above MOMENT\_FT ($0.39$), CHARM ($0.50$) and Time-RCD ($0.52$), below TSPulse\_FT ($0.55$), and $0.06$ below the leaderboard's best entry, Time-RCD\,+\,MAFT ($0.59$).

Pretraining helps on this benchmark, and a detector without it starts at a disadvantage. Within the untrained regime, though, the pipeline dominates: a ten-feature pool with a median/MAD score already clears every untrained baseline, which leaves the question of what stops it from closing the remaining gap. Sections~\ref{sec:knobs} and~\ref{sec:adaptive} answer that the pool is what stops it, and that closing the gap is a matter of writing better features.

\begin{table}[t]
\centering
\setlength{\tabcolsep}{4pt}
\begin{tabular}{llcc}
\toprule
Method & Family & Pretrain. & VUS-PR \\
\midrule
\multicolumn{4}{l}{\emph{Pretrained on time series}} \\
Time-RCD\,+\,MAFT$^{\dagger}$ & foundation & \checkmark & 0.59 \\
TSPulse\_FT$^{\dagger}$       & foundation & \checkmark & 0.55 \\
Time-RCD$^{\dagger}$          & foundation & \checkmark & 0.52 \\
CHARM$^{\dagger}$             & foundation & \checkmark & 0.50 \\
MOMENT\_FT$^{\dagger}$        & foundation & \checkmark & 0.39 \\
\midrule
\multicolumn{4}{l}{\emph{No time-series pretraining}} \\
\textbf{Union pool (ours)}    & statistical & LLM$^{\ddagger}$ & \textbf{0.588} \\
\textbf{Hand-crafted (ours)}  & statistical & --- & \textbf{0.529} \\
StreamVAE$^{\dagger}$         & neural      & --- & 0.45 \\
MMPAD$^{\dagger}$             & statistical & --- & 0.44 \\
Sub\_PCA$^{\dagger}$          & statistical & --- & 0.42 \\
MatrixProfile$^{\dagger}$     & distance    & --- & 0.35 \\
IForest$^{\dagger}$           & isolation   & --- & 0.30 \\
\bottomrule
\end{tabular}
\caption{TSB-AD-U, per-series mean VUS-PR over the 350 evaluation series. $^{\dagger}$as published on the leaderboard; \checkmark marks pretraining on time series. $^{\ddagger}$The union pool's candidate features are written offline by a general-purpose language model, not a time-series foundation model, while its detector is untrained and transductive. Our two rows are the hand-crafted pool (Section~\ref{sec:knobs}) and the union with the generated pool (Table~\ref{tab:generated}, best generator, seed-averaged); the hand-crafted pipeline leads the non-pretrained baselines by $8$ points and the union by $14$, and the union comes close to the leaderboard's best entry.}
\label{tab:main}
\end{table}

\subsection{Selection, aggregation, and the candidate pool}
\label{sec:knobs}

\begin{figure}[t]
\centering
\includegraphics[width=0.85\columnwidth]{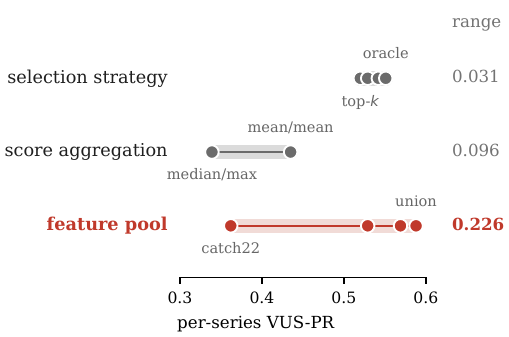}
\caption{Effect size of the three design choices on one axis. Each bar spans the configurations we ran for that choice, dots are individual configurations, and only the two endpoints are named; the number at the right is the range. Selection strategy moves per-series VUS-PR by $0.031$ even when a hindsight oracle is allowed, and score aggregation by $0.096$, while the candidate feature pool moves it by $0.226$. The aggregation sweep runs on the full pool without selection, so that row sits at a lower absolute level than the other two.}
\label{fig:knobs}
\end{figure}

The pipeline exposes three choices and no others: how a feature subset is selected, how the per-feature robust scores are collapsed into a per-point score, and which features are candidates in the first place. We vary each in turn under otherwise identical conditions and measure how far the per-series score moves (Figure~\ref{fig:knobs}). A practitioner would reach first for the selection rule and the aggregation; we show that the candidate pool matters more than either.

\paragraph{Selection strategy is second-order.}
Holding the hand-crafted pool fixed, we vary the selection strategy. Figure~\ref{fig:knobs} (top) shows the three strategies for which we compute full VUS-PR (greedy forward, top-$k$, and mRMR), each selecting on the tuning split and scored on evaluation, alongside a per-domain oracle that picks, in hindsight, the best of the three for each domain. Across all three strategies and the oracle the score spans $0.520$ to $0.551$, a range of $0.031$, and the tuning-selected configuration is within two points of the hindsight oracle. Tuning selected greedy forward in every domain, yielding compact subsets of two to five features. It is the choice \emph{among} strategies that is second-order, not selection itself. Scoring the full ten-feature pool with no selection at all gives $0.435$ (Table~\ref{tab:agg}), so selecting per domain is worth $0.094$.

\paragraph{Score aggregation is second-order too.}
The pipeline collapses the per-feature robust scores $|Z|$ to one value per window (feature aggregation) and then combines overlapping windows into per-point scores (point aggregation). We sweep the full $9 \times 2$ grid on the hand-crafted pool with no selection, so that the sweep is not entangled with the selection step (Table~\ref{tab:agg}). Mean/mean ranks first and is the configuration used for every result in the paper, so all comparisons elsewhere are on matched aggregation.

The grid spans $0.096$, wider than the selection range, but the spread sits almost entirely in the degenerate corners. Averaging over the windows that cover a point beats taking their maximum at all nine feature aggregations, and among the eight non-median feature aggregations under mean point aggregation the spread is only $0.021$. The ordering is interpretable: the score is highest when evidence is pooled across features and decays monotonically as the aggregation moves toward order statistics that rest on fewer and fewer of them ($q_{75} \rightarrow q_{90} \rightarrow q_{95} \rightarrow q_{99} \rightarrow \max$).

\begin{table}[t]
\centering
\setlength{\tabcolsep}{6pt}
\begin{tabular}{lcc}
\toprule
 & \multicolumn{2}{c}{point aggregation} \\
\cmidrule(lr){2-3}
feature aggregation & mean & max \\
\midrule
mean      & \textbf{0.435} & 0.389 \\
$q_{75}$  & 0.433 & 0.389 \\
top-3     & 0.430 & 0.384 \\
$q_{90}$  & 0.428 & 0.384 \\
top-2     & 0.424 & 0.380 \\
$q_{95}$  & 0.424 & 0.380 \\
$q_{99}$  & 0.416 & 0.372 \\
max       & 0.414 & 0.366 \\
median    & 0.388 & 0.339 \\
\bottomrule
\end{tabular}
\caption{Score-aggregation ablation (per-series VUS-PR), full hand-crafted pool with no feature selection, so the absolute level is below the selected pipeline ($0.529$) throughout. Rows are ordered by the mean point-aggregation column. Mean point aggregation dominates max at every one of the nine feature aggregations, and mean/mean is best overall; it is the setting used for every other result in the paper.}
\label{tab:agg}
\end{table}

\paragraph{The pool is the ceiling.}
Against those two, we now vary the candidate pool and hold everything else fixed, and we test the standard response to a weak pool, a larger curated library. Table~\ref{tab:pools} reports four pools: catch22, the canonical off-the-shelf library of 22 generic time-series features~\cite{catch22}; our hand-crafted pool of ten statistics; the adaptively generated pool of the next section (\texttt{gemini-3.5-flash}); and the union of the last two. Every pool runs through the identical detector, window, aggregation, and selection protocol. The one difference is that greedy forward selection is too costly for the two large pools, so catch22 and the union pool select with top-$k$ and mRMR only. This does not manufacture either result. On the hand-crafted pool mRMR is the strongest of the three strategies and greedy the weakest (Figure~\ref{fig:knobs}).

\begin{table}[t]
\centering
\setlength{\tabcolsep}{5pt}
\begin{tabular}{lcc}
\toprule
Feature pool & \#feat. & VUS-PR \\
\midrule
Union (generated $\cup$ hand) & 20 & \textbf{0.588 {\footnotesize$\pm$0.016}} \\
Generated (\texttt{gemini-3.5}) & 10 & 0.569 {\footnotesize$\pm$0.025} \\
Hand-crafted statistics & 10 & 0.529 \\
catch22 (generic library) & 22 & 0.362 \\
\bottomrule
\end{tabular}
\caption{The pool is the ceiling (per-series VUS-PR). Same detector, window, aggregation, and per-domain selection; only the candidate pool changes. The generic catch22 library is the weakest despite being the largest; the adaptively generated pool beats the hand-crafted one; their union is best. The generated and union rows are the mean $\pm$ standard deviation over four generation seeds.}
\label{tab:pools}
\end{table}

The pool alone moves per-series VUS-PR from $0.362$ (catch22) to $0.588$ (union), a range seven times the selection range and more than twice the full aggregation grid, and the aggregation grid earns even that width from settings no one would choose. Feature generation, not feature selection or score aggregation, is the dominant factor.

The larger static library is the weakest pool: catch22 has more than twice the features of our hand-crafted pool yet scores well below it, so enlarging a generic vocabulary is the wrong axis. catch22 is built for normalised series and largely omits location, which is what several domains' level-shift anomalies require. The adaptively generated pool, by contrast, already beats the hand-crafted one, and the union of the two is best, which the next two sections examine across generators.

\subsection{Adaptive feature discovery}
\label{sec:adaptive}

We now replace the static pool with an automatically discovered one and hold everything else fixed. Table~\ref{tab:generated} reports the two feature sources of Section~\ref{sec:generation}: \texttt{llm}, which selects over the generated pool alone, and \texttt{union}, which selects jointly over the generated and hand-crafted pools. Each entry is the mean over four seeds.

The stronger generators are competitive on their own. Both Gemini pools match or beat the hand-crafted pipeline (per-series $0.529$) with no manual feature engineering, \texttt{gemini-3.5-flash} reaching $0.569$; the weaker \texttt{claude-sonnet-5} pool trails at $0.497$, so ``competitive alone'' holds for the better generators but not uniformly. Discovery can substitute for the manual design of the pool, but it is not reliable on its own. The complementarity is the more robust result. The cleanest evidence avoids the series-correlation problem entirely: within a seed the \texttt{union} pool is exactly that seed's generated pool plus the hand-crafted features, so \texttt{union} $-$ \texttt{llm} is an exact paired difference. Across three generators and four seeds, \texttt{union} wins all twelve pairs, by margins from $0.006$ to $0.061$ (two-sided sign test, $p = 4.9 \times 10^{-4}$). The largest per-seed margins belong to \texttt{claude-sonnet-5}, the weakest generator, which shows the blind-spot mechanism at the level of individual runs. The strongest union configuration (\texttt{gemini-3.5-flash}, per-series $0.588$) exceeds both the hand-crafted pipeline and every generated pool in isolation.

\begin{table}[t]
\centering
\setlength{\tabcolsep}{6pt}
\begin{tabular}{lcc}
\toprule
Generator & \texttt{llm} & \texttt{union} \\
\midrule
gemini-3.5-flash      & 0.569 {\footnotesize$\pm$0.025} & \textbf{0.588 {\footnotesize$\pm$0.016}} \\
gemini-3.1-flash-lite & 0.540 {\footnotesize$\pm$0.018} & 0.564 {\footnotesize$\pm$0.025} \\
claude-sonnet-5       & 0.497 {\footnotesize$\pm$0.024} & 0.551 {\footnotesize$\pm$0.028} \\
\bottomrule
\end{tabular}
\caption{Generated pools (per-series VUS-PR), mean $\pm$ standard deviation over four generation seeds. \texttt{llm} selects over the generated pool alone; \texttt{union} over the generated and hand-crafted pools jointly. \texttt{union} beats \texttt{llm} for every generator, and the best \texttt{union} beats the hand-crafted pipeline ($0.529$).}
\label{tab:generated}
\end{table}

\subsection{Per-domain analysis}
\label{sec:perdomain}

Table~\ref{tab:perdomain} decomposes the best generator by domain against the hand-crafted pipeline. Neither pool dominates: the hand-crafted one leads on finance, medical, and traffic, the generated one on facility and web services, and the two differ by more than $0.27$ on human activity in the generated pool's favour. The union does more than pick the better parent. It exceeds \emph{both} pools on environment, human activity, sensor, and synthetic, so the selected subset draws features from each source rather than collapsing onto the stronger one, and where it trails the better parent it trails by $0.03$ or less, except on traffic, the smallest domain at five series.

Human activity is the clearest case: the hand-crafted pool is invariant to the level shifts that the generated code targets (Listing~\ref{lst:humanactivity}), and scores $0.238$ against $0.513$ for the generated pool.

\begin{table}[t]
\centering
\setlength{\tabcolsep}{3pt}
\begin{tabular}{lcccc}
\toprule
Domain & $n$ & Hand & \texttt{llm} & \texttt{union} \\
\midrule
environment    & 18 & 0.360 & 0.348 & \textbf{0.408} \\
facility       & 87 & 0.603 & \textbf{0.673} & 0.656 \\
finance        & 8  & \textbf{0.768} & 0.745 & 0.743 \\
human activity & 43 & 0.238 & 0.513 & \textbf{0.528} \\
medical        & 47 & \textbf{0.593} & 0.540 & 0.590 \\
sensor         & 39 & 0.686 & 0.643 & \textbf{0.756} \\
synthetic      & 39 & 0.483 & 0.471 & \textbf{0.502} \\
traffic        & 5  & \textbf{0.440} & 0.394 & 0.394 \\
web services   & 64 & 0.534 & \textbf{0.554} & 0.530 \\
\midrule
per-series     & 350 & 0.529 & 0.569 & \textbf{0.588} \\
per-domain     & 9   & 0.523 & 0.543 & \textbf{0.567} \\
\bottomrule
\end{tabular}
\caption{Per-domain VUS-PR for the best generator (\texttt{gemini-3.5-flash}, seed-averaged). ``Hand'' is the hand-crafted pipeline, \texttt{llm} the generated pool, \texttt{union} the two pools selected over jointly; best per row in bold. Neither source wins everywhere, and the union exceeds both on four domains, which is why it leads overall.}
\label{tab:perdomain}
\end{table}

\section{Discussion and Conclusion}

Feature-based anomaly detection has largely relied on static, expert-curated libraries and on searching well within them. In our setting the search is close to exhausted: three selection strategies and a hindsight oracle lie within $0.031$ of one another, and the defensible aggregation settings within $0.021$, while changing the candidate pool moves the score by $0.226$. What limits this pipeline is therefore the pool it is handed rather than the search over it, and the one larger generic library we tried did not help: catch22 is the weakest of the four pools we run.

Task-conditioned discovery addresses that constraint directly. A static library encodes prior expert knowledge, a discovered pool encodes hypotheses conditioned on the target domain, and the two fail on disjoint subsets of domains. Selecting over their union exceeds both constituents on four of nine domains and improves on the generated pool in all twelve generator-seed pairs, with the largest gains where the generated pool is weakest. Augmenting an expert pool with discovered candidates is therefore low-risk and does not depend on any individual generator: strong generation lifts the result, and weak generation leaves the expert features in the candidate set.

The resulting detector has no trained parameters and scores each series transductively. In its hand-crafted form it uses no pretraining at all and still exceeds every non-pretrained entry on the TSB-AD-U leaderboard by a wide margin; unioning in a language-model-written pool matches the best pretrained entry, using a pretrained model to design features offline rather than to score.

\paragraph{Limitations.}
The complementarity is consistent across seeds but concentrated in the blind-spot domains, and nine domains afford little power for per-generator claims. The generator comparison covers three models and four seeds each rather than an exhaustive sweep. The leaderboard entries are tuned by their own authors under the benchmark's protocol rather than ours which requires less tuning, so we make claims at the level of method families and do not claim to exceed the best pretrained entry, only to reach it.

\paragraph{Future work.}
Two natural extensions remain. Applying the protocol to the multivariate TSB-AD track would test whether the findings extend beyond the univariate setting. An iterative loop that returns the tuning score of a pool to the generator and requests a revision would establish whether a single generation round leaves gains unrealised.

\bibliographystyle{plainnat}
\bibliography{references}

@article{paparrizos2022tsb,
  title={Tsb-uad: an end-to-end benchmark suite for univariate time-series anomaly detection},
  author={Paparrizos, John and Kang, Yuhao and Boniol, Paul and Tsay, Ruey S and Palpanas, Themis and Franklin, Michael J},
  journal={Proceedings of the VLDB Endowment},
  volume={15},
  number={8},
  pages={1697--1711},
  year={2022},
  publisher={VLDB Endowment}
}

@misc{ClaudeSonnet52026,
  author       = {Anthropic},
  title        = {Claude Sonnet 5},
  year         = {2026},
  howpublished = {Large language model},
  url          = {https://claude.ai}
}

@misc{gemini35flash,
  author = {{Google DeepMind}},
  title = {Gemini 3.5 Flash: Model Card},
  year = {2026},
  url = {https://deepmind.google/models/model-cards/gemini-3-5-flash/}
}

@techreport{gemini31flashlite,
  title     = {Gemini 3.1 Flash-Lite: Model Card},
  author    = {Google DeepMind},
  year      = {2026},
  month     = {March},
  url       = {https://deepmind.google/models/model-cards/gemini-3-1-flash-lite/}
}

@article{paparrizos2022volume,
  title={{Volume Under the Surface: A New Accuracy Evaluation Measure for Time-Series Anomaly Detection}},
  author={Paparrizos, John and Boniol, Paul and Palpanas, Themis and Tsay, Ruey S and Elmore, Aaron and Franklin, Michael J},
  journal={Proceedings of the VLDB Endowment},
  volume={15},
  number={11},
  pages={2774--2787},
  year={2022},
  publisher={VLDB Endowment}
}

@misc{streamvae,
      title={STREAM-VAE: Dual-Path Routing for Slow and Fast Dynamics in Vehicle Telemetry Anomaly Detection}, 
      author={Kadir-Kaan Özer and René Ebeling and Markus Enzweiler},
      year={2026},
      eprint={2511.15339},
      archivePrefix={arXiv},
      primaryClass={cs.LG},
      url={https://arxiv.org/abs/2511.15339}, 
}

@inproceedings{
gruver2023large,
title={Large Language Models Are Zero-Shot Time Series Forecasters},
author={Nate Gruver and Marc Anton Finzi and Shikai Qiu and Andrew Gordon Wilson},
booktitle={Thirty-seventh Conference on Neural Information Processing Systems},
year={2023},
url={https://openreview.net/forum?id=md68e8iZK1}
}

@article{MAD,
title = {Detecting outliers: Do not use standard deviation around the mean, use absolute deviation around the median},
journal = {Journal of Experimental Social Psychology},
volume = {49},
number = {4},
pages = {764-766},
year = {2013},
issn = {0022-1031},
doi = {https://doi.org/10.1016/j.jesp.2013.03.013},
url = {https://www.sciencedirect.com/science/article/pii/S0022103113000668},
author = {Christophe Leys and Christophe Ley and Olivier Klein and Philippe Bernard and Laurent Licata}}

@inproceedings{jin2023time,
  title={{Time-LLM}: Time series forecasting by reprogramming large language models},
  author={Jin, Ming and Wang, Shiyu and Ma, Lintao and Chu, Zhixuan and Zhang, James Y and Shi, Xiaoming and Chen, Pin-Yu and Liang, Yuxuan and Li, Yuan-Fang and Pan, Shirui and Wen, Qingsong},
  booktitle={International Conference on Learning Representations (ICLR)},
  year={2024}
}

@inproceedings{
timercd,
title={Towards Foundation Models for Zero-Shot Time Series Anomaly Detection: Leveraging Synthetic Data and Relative Context Discrepancy},
author={Tian Lan and Hao Duong Le and Jinbo Li and Wenjun He and Meng Wang and Chenghao Liu and Chen Zhang},
booktitle={Forty-third International Conference on Machine Learning},
year={2026},
url={https://openreview.net/forum?id=yXqnyIvGAy}
}

@inproceedings{liu2024elephant,
  title={The Elephant in the Room: Towards A Reliable Time-Series Anomaly Detection Benchmark},
  author={Liu, Qinghua and Paparrizos, John},
  booktitle={NeurIPS 2024},
  year={2024}
}

@misc{tspulse,
      title={TSPulse: Tiny Pre-Trained Models with Disentangled Representations for Rapid Time-Series Analysis}, 
      author={Vijay Ekambaram and Subodh Kumar and Arindam Jati and Sumanta Mukherjee and Tomoya Sakai and Pankaj Dayama and Wesley M. Gifford and Jayant Kalagnanam},
      year={2026},
      eprint={2505.13033},
      archivePrefix={arXiv},
      primaryClass={cs.LG},
      url={https://arxiv.org/abs/2505.13033}, 
}

@article{catch22,
  title = {catch22: CAnonical Time-series CHaracteristics: Selected through highly comparative time-series analysis},
  volume = {33},
  ISSN = {1573-756X},
  url = {http://dx.doi.org/10.1007/s10618-019-00647-x},
  DOI = {10.1007/s10618-019-00647-x},
  number = {6},
  journal = {Data Mining and Knowledge Discovery},
  publisher = {Springer Science and Business Media LLC},
  author = {Lubba,  Carl H. and Sethi,  Sarab S. and Knaute,  Philip and Schultz,  Simon R. and Fulcher,  Ben D. and Jones,  Nick S.},
  year = {2019},
  month = Aug,
  pages = {1821–1852}
}

@article{tsfresh,
title = {Time Series FeatuRe Extraction on basis of Scalable Hypothesis tests (tsfresh – A Python package)},
journal = {Neurocomputing},
volume = {307},
pages = {72-77},
year = {2018},
issn = {0925-2312},
doi = {https://doi.org/10.1016/j.neucom.2018.03.067},
url = {https://www.sciencedirect.com/science/article/pii/S0925231218304843},
author = {Maximilian Christ and Nils Braun and Julius Neuffer and Andreas W. Kempa-Liehr},
}

@misc{CAAFE,
      title={LLMs for Semi-Automated Data Science: Introducing CAAFE for Context-Aware Automated Feature Engineering}, 
      author={Noah Hollmann and Samuel Müller and Frank Hutter},
      year={2023},
      eprint={2305.03403},
      archivePrefix={arXiv},
      primaryClass={cs.AI}
}

@INPROCEEDINGS{iforest,
  author={Liu, Fei Tony and Ting, Kai Ming and Zhou, Zhi-Hua},
  booktitle={2008 Eighth IEEE International Conference on Data Mining}, 
  title={Isolation Forest}, 
  year={2008},
  volume={},
  number={},
  pages={413-422},
  doi={10.1109/ICDM.2008.17}}

@misc{xu2026treestructuredsynergylargelanguage,
      title={Tree-Structured Synergy of Large Language Models and Bayesian Optimization for Efficient CASH}, 
      author={Beicheng Xu and Weitong Qian and Lingching Tung and Yupeng Lu and Bin Cui},
      year={2026},
      eprint={2601.12355},
      archivePrefix={arXiv},
      primaryClass={cs.LG},
      url={https://arxiv.org/abs/2601.12355}, 
}

@article{
ansari2024chronos,
title={Chronos: Learning the Language of Time Series},
author={Abdul Fatir Ansari and Lorenzo Stella and Ali Caner Turkmen and Xiyuan Zhang and Pedro Mercado and Huibin Shen and Oleksandr Shchur and Syama Sundar Rangapuram and Sebastian Pineda Arango and Shubham Kapoor and Jasper Zschiegner and Danielle C. Maddix and Hao Wang and Michael W. Mahoney and Kari Torkkola and Andrew Gordon Wilson and Michael Bohlke-Schneider and Bernie Wang},
journal={Transactions on Machine Learning Research},
issn={2835-8856},
year={2024},
url={https://openreview.net/forum?id=gerNCVqqtR},
note={Expert Certification}
}

@inproceedings{goswami2024moment,
  title={MOMENT: A Family of Open Time-series Foundation Models},
  author={Mononito Goswami and Konrad Szafer and Arjun Choudhry and Yifu Cai and Shuo Li and Artur Dubrawski},
  booktitle={International Conference on Machine Learning},
  year={2024}
}

@misc{CHARM,
      title={Time to Embed: Unlocking Foundation Models for Time Series with Channel Descriptions}, 
      author={Utsav Dutta and Sina Khoshfetrat Pakazad and Henrik Ohlsson},
      year={2025},
      eprint={2505.14543},
      archivePrefix={arXiv},
      primaryClass={cs.LG},
      url={https://arxiv.org/abs/2505.14543}, 
}

@inproceedings{matrixprofile,
author = {Yeh, Chin-Chia Michael and Zhu, Yan and Ulanova, Liudmila and Begum, Nurjahan and Ding, Yifei and Dau, Anh and Silva, Diego and Mueen, Abdullah and Keogh, Eamonn},
year = {2016},
month = {12},
pages = {1317-1322},
title = {Matrix Profile I: All Pairs Similarity Joins for Time Series: A Unifying View That Includes Motifs, Discords and Shapelets},
doi = {10.1109/ICDM.2016.0179}
}

@Article{FunSearch2023,
  author  = {Romera-Paredes, Bernardino and Barekatain, Mohammadamin and Novikov, Alexander and Balog, Matej and Kumar, M. Pawan and Dupont, Emilien and Ruiz, Francisco J. R. and Ellenberg, Jordan and Wang, Pengming and Fawzi, Omar and Kohli, Pushmeet and Fawzi, Alhussein},
  journal = {Nature},
  title   = {Mathematical discoveries from program search with large language models},
  year    = {2023},
  doi     = {10.1038/s41586-023-06924-6}
}

@misc{mmpad,
      title={Matrix Profile for Time-Series Anomaly Detection: A Reproducible Open-Source Benchmark on TSB-AD}, 
      author={Chin-Chia Michael Yeh},
      year={2026},
      eprint={2604.02445},
      archivePrefix={arXiv},
      primaryClass={cs.LG},
      url={https://arxiv.org/abs/2604.02445}, 
}

\clearpage
\appendix

\section*{Appendix}
\addcontentsline{toc}{section}{Appendix}

\noindent
This appendix records the experimental setup and the full results
behind the reported aggregates: the dataset splits and evaluation metric, the configuration and
inference cost of the language-model generators, a breakdown of every pool by
selection strategy, the selected feature subsets, and the seed-level scores with
their statistical analysis.

\section{Dataset and Splits}
\label{app:data}

Experiments are run on TSB-AD-U, the univariate track of
TSB-AD~\cite{liu2024elephant}. The benchmark's fixed Tuning/Eval partition is
adopted without modification: 48 tuning series and 350 evaluation series,
grouped into nine domains by the benchmark's filename convention.
Table~\ref{tab:splits} gives the per-domain counts. Two domains, traffic and
finance, are small on both sides of the split, which limits the precision of any
domain-level statement about them.

The nine domains are the benchmark's own per-series labels. Each TSB-AD-U series is distributed under a filename that
encodes both its source dataset and its domain, as in
\texttt{001\_NAB\_id\_1\_} \texttt{Facility\_tr\_1007\_1st\_2014.csv}; we read the domain
token off the filename and normalise it, and no series in either split falls
outside the nine labels.

The only per-series preprocessing is extraction of channel 0 as a float array.
No normalisation, resampling, detrending, or imputation is applied at any point,
and no series is excluded: all 350 evaluation series are scored, for every
feature source and every seed.

\begin{table}[H]
\centering
\setlength{\tabcolsep}{6pt}
\begin{tabular}{lcc}
\toprule
Domain & Tuning & Eval \\
\midrule
environment    & 2  & 18 \\
facility       & 9  & 87 \\
finance        & 2  & 8 \\
human activity & 3  & 43 \\
medical        & 7  & 47 \\
sensor         & 5  & 39 \\
synthetic      & 6  & 39 \\
traffic        & 1  & 5 \\
web services   & 13 & 64 \\
\midrule
Total          & 48 & 350 \\
\bottomrule
\end{tabular}
\caption{TSB-AD-U series counts per domain, as distributed by the benchmark.}
\label{tab:splits}
\end{table}

\paragraph{Evaluation metric.}
All scores are VUS-PR~\cite{paparrizos2022volume}, computed by the benchmark's
own evaluation code so that these numbers and the published leaderboard values
are produced by identical software. VUS-PR integrates the precision-recall curve
over a range of buffer widths around each labelled anomalous region, which makes
it threshold-free and tolerant of small localisation errors. Following the
TSB-AD protocol, the buffer parameter is set per series to that series'
estimated dominant period, obtained by the benchmark's unsupervised period
estimator on the raw values.

Two aggregates are used throughout. \emph{Per-series} is the mean VUS-PR over
all 350 evaluation series, equivalently the domain means weighted by their
evaluation counts; this is the aggregate the benchmark reports on its
leaderboard. \emph{Per-domain} is the unweighted mean of the nine domain means.
Unless stated otherwise, figures quoted in the text are per-series.

\section{Pipeline Hyperparameters}
\label{app:hparams}

Table~\ref{tab:hparams} lists every hyperparameter of the detection pipeline,
the values tried, and the final setting. The pipeline has no learned parameters:
fit and score both reduce to per-column medians on the series being scored.

Labels are consulted in exactly one place, the per-domain feature-selection step, and that
step runs strictly on the tuning split; the 350 evaluation series are scored
once, after all design choices are frozen. The window is not a fitted quantity:
$w$ is the series' own unsupervised dominant-period estimate and
$s = \lfloor w/2 \rfloor$, so there is nothing to tune. Selection optimises a
cheap AUC-PR proxy to keep the search affordable, while every reported
evaluation score uses full VUS-PR.

Only three things in the table are searched at all: the two score-aggregation
choices, swept once on the hand-crafted pool without selection, and the feature
subset itself, chosen per domain on the tuning split. Everything under
\emph{Windowing} and the detector's estimator are fixed a priori and never
varied. One practical restriction applies to the search: greedy forward is
$O(d^2)$ in objective evaluations, so on the two pools of twenty features or
more, catch22 and the union pool, only top-$k$ and mRMR are run.

\begin{table}[h]
\centering
\setlength{\tabcolsep}{4pt}
\footnotesize
\begin{tabular}{@{}p{2.3cm}p{2.9cm}p{2.3cm}@{}}
\toprule
Hyper\-parameter & Values tried & Final \\
\midrule
\multicolumn{3}{@{}l}{\emph{Windowing} --- fixed a priori, not searched} \\
window length $w$   & --- & per-series dominant period \\
stride $s$          & --- & $\lfloor w/2 \rfloor$ \\
period floor / cap  & --- & 4 / series length \\
\midrule
\multicolumn{3}{@{}l}{\emph{Detector}} \\
location / scale    & --- & median / MAD \\
zero-MAD substitute & --- & $1.0$ \\
feature aggregation & 9: mean, median, max, $q_{75/90/95/99}$, top-2, top-3 & mean \\
point aggregation   & \{mean, max\} & mean \\
\midrule
\multicolumn{3}{@{}l}{\emph{Feature selection} --- tuning split only} \\
strategy            & \{greedy forward, top-$k$, mRMR\} & per domain \\
objective           & --- & mean AUC-PR on tuning series \\
greedy stopping     & --- & no candidate improves \\
top-$k$ grid        & $k \in \{1,\dots,d\}$ & per domain \\
mRMR $\lambda$ grid & \{0, 0.25, 0.5, 1.0\} & per domain \\
mRMR redundancy     & --- & $|{\rm corr}|$ on tuning windows \\
\midrule
\multicolumn{3}{@{}l}{\emph{Candidate pool} --- the experimental variable} \\
pool size $d$       & 10 hand, 22 catch22, 10 llm, 20 union & varied \\
\bottomrule
\end{tabular}
\caption{Pipeline hyperparameters. A dash under \emph{Values tried} means the
setting was fixed a priori and never varied; only the two aggregation choices
and the feature subset are searched at all, both on the tuning split. The
detector has no learned parameters.}
\label{tab:hparams}
\end{table}

\paragraph{The hand-crafted pool.}
The ten hand-crafted candidates are: standard deviation, interquartile range,
skewness, excess kurtosis, mean absolute first difference, mean absolute second
difference, mean-crossing count, lag-1 autocorrelation, lag-2 autocorrelation,
and a 10-bin histogram entropy. Each is a pure function of a single window
returning one finite scalar; non-finite returns are replaced by zero at extraction time.

\section{Language-Model Generator Configuration}
\label{app:llm}

Table~\ref{tab:llmhparams} lists the generation settings, held fixed across the
three generators except where a provider does not expose the parameter. The
prompt is reproduced in Section~\ref{app:prompt}.

Each domain gets one pool, generated from ten rendered example windows,  five
containing a labelled anomaly, five not, drawn from that domain's tuning
series. The seed controls only which windows are sampled; the instruction, the
rendering, and everything downstream are identical across domains, generators,
and seeds. The returned code is parsed from a single fenced block, executed in a
restricted namespace with numpy available, and each function is checked on a
small set of probe windows covering the edge cases named in the prompt (very
short, constant, and large-magnitude windows). Functions that raise or that
always return non-finite values are discarded, and a run yielding fewer than ten
valid features is retried rather than used. Across all 108 (generator, seed,
domain) runs the accepted pool contains exactly ten features, so no result below
rests on a partially populated pool.

\begin{table}[h]
\centering
\setlength{\tabcolsep}{4pt}
\small
\begin{tabular}{@{}lp{4.5cm}@{}}
\toprule
Setting & Value \\
\midrule
Generators & \texttt{gemini-3.5-flash}~\cite{gemini35flash}, \\
           & \texttt{gemini-3.1-flash-lite}~\cite{gemini31flashlite}, \\
           & \texttt{claude-sonnet-5}~\cite{ClaudeSonnet52026} \\
Features requested & 10 per domain \\
Accepted pool size & 10 (a run yielding fewer is retried) \\
Example windows & 5 normal $+$ 5 anomalous per domain \\
Window rendering & line plot, shared $y$-axis across all \\
                 & ten plots \\
Seeds & 0, 1, 2, 3  \\
Temperature & 0.2 (Gemini); not exposed by the \\
            & Anthropic endpoint for this model \\
Max output tokens & 8192 \\
Reasoning & Gemini: provider default; \\
          & Claude: adaptive thinking, effort \texttt{high} \\
\bottomrule
\end{tabular}
\caption{Generation settings. Only the attached plots change between domains and
seeds; the instruction is byte-identical throughout.}
\label{tab:llmhparams}
\end{table}

\section{Inference Cost of the Generators}
\label{app:cost}

Generation is a one-off, offline cost, and the natural unit is a single domain:
one API call produces one domain's ten-function pool. The detector that then
runs on the 350 evaluation series never contacts a language model.

\begin{table}[h]
\centering
\setlength{\tabcolsep}{3.5pt}
\small
\begin{tabular}{@{}lrrrrr@{}}
\toprule
Generator & Input & Output & Total & USD & s \\
\midrule
gemini-3.5-flash      & 11{,}180 & 546   & 17{,}075 & 0.070 & 34.9 \\
gemini-3.1-flash-lite & 11{,}180 & 512   & 11{,}692 & 0.004 & 2.7 \\
claude-sonnet-5       & 1{,}767  & 1{,}744 & 3{,}511 & 0.031 & 14.0 \\
\bottomrule
\end{tabular}
\caption{Average tokens, list-price cost and latency for one generation call,
which produces one domain's pool, measured from the providers' usage telemetry.
Prices are the providers' published current rates per million input/output
tokens: $1.50$/$9.00$, $0.25$/$1.50$ and $3.00$/$15.00$ respectively.}
\label{tab:tokens}
\end{table}

Input dominates, and differs by roughly $6\times$ between providers because of
how each tokenises the ten plots. Output is small in every case, being ten short
numpy functions. For \texttt{gemini-3.5-flash} the total exceeds input $+$
output because reasoning tokens are billed separately, a mean of $5{,}349$ per
call; for \texttt{claude-sonnet-5} they are already inside the output count. Equipping a domain with a
generated pool costs between half a cent and seven cents, once.

\section{Baselines}
\label{app:baselines}

Every baseline figure is taken as published on the public TSB-AD-U leaderboard.
No baseline was trained, tuned, or re-run here, so no hyperparameters are
reported for them: the settings are the original authors' under the benchmark's
protocol, and the leaderboard and the respective papers are the reference for
those details. This applies to MOMENT\_FT~\cite{goswami2024moment}, Time-RCD and
Time-RCD\,+\,MAFT~\cite{timercd}, CHARM~\cite{CHARM}, TSPulse\_FT~\cite{tspulse},
StreamVAE~\cite{streamvae}, MMPAD~\cite{mmpad}, Sub\_PCA~\cite{liu2024elephant},
MatrixProfile~\cite{matrixprofile}, and IForest~\cite{iforest}.

\section{Results by Selection Strategy}
\label{app:strategy}

Table~\ref{tab:bystrategy} reports per-series VUS-PR for each candidate pool
under each selection strategy that is affordable on it. Entries for the
generated (\texttt{llm}) and union pools are means over the four generation
seeds; the hand-crafted row is a single deterministic run. Greedy forward is
$O(d^2)$ in objective evaluations and is not run on the 20-feature union pool.

\begin{table}[h]
\centering
\setlength{\tabcolsep}{4pt}
\small
\begin{tabular}{@{}llccc@{}}
\toprule
Generator & Pool & greedy & top-$k$ & mRMR \\
\midrule
gemini-3.5-flash & \texttt{llm}   & 0.572 & 0.576 & 0.573 \\
                 & \texttt{union} & ---   & 0.571 & \textbf{0.588} \\
\addlinespace
gemini-3.1-      & \texttt{llm}   & 0.541 & 0.543 & 0.540 \\
flash-lite       & \texttt{union} & ---   & 0.538 & \textbf{0.564} \\
\addlinespace
claude-sonnet-5  & \texttt{llm}   & 0.491 & 0.496 & 0.495 \\
                 & \texttt{union} & ---   & 0.526 & \textbf{0.551} \\
\midrule
hand-crafted     & \texttt{hand}  & 0.529 & 0.521 & 0.543 \\
\bottomrule
\end{tabular}
\caption{Per-series VUS-PR by selection strategy. Generated and union entries are
seed-averaged over four generation seeds; the hand-crafted entry is a single
deterministic run.}
\label{tab:bystrategy}
\end{table}

\paragraph{Strategy has little effect on a fixed pool.}
Within the \texttt{llm} row of any generator the three strategies lie within
$0.005$ of one another: $0.572$/$0.576$/$0.573$ for
\texttt{gemini-3.5-flash}, $0.541$/$0.543$/$0.540$ for
\texttt{gemini-3.1-flash-lite}, and $0.491$/$0.496$/$0.495$ for
\texttt{claude-sonnet-5}. The hand-crafted pool spans a slightly wider
$0.521$--$0.543$.

\paragraph{The pool matters more than the strategy.}
The spread across strategies within a row is at most $0.025$ (union,
\texttt{claude-sonnet-5}) and typically under $0.005$. The spread across pools at
a fixed strategy is several times larger: under mRMR alone the score runs from
$0.495$ (\texttt{claude-sonnet-5} generated) to $0.588$
(\texttt{gemini-3.5-flash} union). The fourth candidate pool, catch22, is not
broken out by strategy in Table~\ref{tab:bystrategy}; at $0.362$ per-series it
sits below every entry there despite being the largest of the four pools at 22
features. The ordering of pools is stable under every strategy.

\paragraph{The union pool's advantage depends on the strategy.}
mRMR on the union pool is the best configuration for all three generators. For
\texttt{claude-sonnet-5} the union pool beats the generated pool under both
available strategies, $0.526$ and $0.551$ against a best of $0.496$. For the two
Gemini generators the union pool wins only under mRMR: under top-$k$ it is level
with or slightly below the generated pool alone, $0.571$ against $0.576$ and
$0.538$ against $0.543$. Enlarging the candidate set therefore pays off when the
selector penalises redundancy, and not otherwise. This is consistent with the union
pool containing near-duplicate features across its two halves, which top-$k$
ignores by construction. The margin is largest for the weakest generator, whose
pool leaves the most for the hand-crafted features to cover.

\section{Selected Feature Subsets}
\label{app:selected}

Table~\ref{tab:selected} gives the subsets greedy forward selects on the
hand-crafted pool, together with the tuning objective that drove the choice and
the resulting evaluation score. Subsets contain two to five of the ten
candidates.

The tuning and evaluation columns are computed on disjoint series and diverge
sharply in the two smallest domains, finance with two tuning series and traffic
with one.

\begin{table}[h]
\centering
\setlength{\tabcolsep}{3pt}
\small
\begin{tabular}{@{}lp{3.5cm}cc@{}}
\toprule
Domain & Selected features & Tun. & Eval \\
\midrule
environment    & skewness, mean abs.\ change & 0.430 & 0.360 \\
facility       & IQR, skewness & 0.751 & 0.604 \\
finance        & std, mean abs.\ change, skewness & 0.212 & 0.768 \\
human activity & zero crossings, autocorr.\ lag-2, skewness & 0.213 & 0.238 \\
medical        & skewness, std, entropy & 0.513 & 0.593 \\
sensor         & std, IQR & 0.732 & 0.686 \\
synthetic      & mean abs.\ change, mean 2nd deriv. & 0.365 & 0.483 \\
traffic        & std, skewness, mean 2nd deriv., autocorr.\ lag-2 & 0.901 & 0.440 \\
web services   & mean abs.\ change, entropy, autocorr.\ lag-2, std, zero crossings & 0.338 & 0.534 \\
\bottomrule
\end{tabular}
\caption{Feature subsets selected by greedy forward on the hand-crafted pool.
``Tun.'' is the tuning objective used to make the choice; ``Eval'' is the full
VUS-PR on that domain's evaluation series. The per-series mean of the Eval
column over all 350 series is $0.529$.}
\label{tab:selected}
\end{table}

The corresponding subsets for the generated and union pools are not tabulated
here, since there are 108 of them and they change with the seed. In aggregate, selected subsets
contain 1--8 features on the generated pool (mean $3.2$) and 1--20 on the union
pool (mean $4.2$), so selection on the union pool draws from both halves rather
than collapsing onto one.

\newpage
\section{Seed-Level Results and Statistical Analysis}
\label{app:seeds}

Pool generation is the only stochastic component, and each (generator, source) cell
is four independent runs, one per generation seed. Every run scores all 350
evaluation series.

Table~\ref{tab:seedlevel} gives the individual per-seed scores and the
within-seed difference between the two pools. Pairing is exact rather than
approximate: the two members of a pair differ only in whether the ten
hand-crafted features were among the candidates, and are evaluated on the same
350 series with the same detector, so the usual concern about correlated series
does not arise.

\begin{table}[H]
\centering
\setlength{\tabcolsep}{5pt}
\small
\begin{tabular}{@{}llccc@{}}
\toprule
Generator & Seed & \texttt{llm} & \texttt{union} & $\Delta$ \\
\midrule
gemini-3.5-flash & 0 & 0.563 & 0.591 & $+0.028$ \\
                 & 1 & 0.589 & 0.599 & $+0.010$ \\
                 & 2 & 0.531 & 0.562 & $+0.031$ \\
                 & 3 & 0.594 & 0.600 & $+0.006$ \\
\addlinespace
gemini-3.1-flash-lite & 0 & 0.547 & 0.583 & $+0.036$ \\
                      & 1 & 0.533 & 0.545 & $+0.012$ \\
                      & 2 & 0.515 & 0.536 & $+0.021$ \\
                      & 3 & 0.564 & 0.594 & $+0.030$ \\
\addlinespace
claude-sonnet-5 & 0 & 0.537 & 0.597 & $+0.061$ \\
                & 1 & 0.476 & 0.537 & $+0.060$ \\
                & 2 & 0.495 & 0.546 & $+0.051$ \\
                & 3 & 0.481 & 0.526 & $+0.046$ \\
\bottomrule
\end{tabular}
\caption{Per-seed per-series VUS-PR over the 350 evaluation series. Within a seed, the \texttt{union} pool is
exactly that seed's generated pool plus the ten hand-crafted features, so
$\Delta$ is an exact paired difference. All twelve differences are positive.}
\label{tab:seedlevel}
\end{table}

\paragraph{Statistical test.}
A two-sided sign test on 12 positives out of 12 gives
$p = 2^{-11} = 4.9\times10^{-4}$, the smallest value attainable at this sample
size; the Wilcoxon signed-rank test agrees ($W = 0$, $p = 4.9\times10^{-4}$). The
mean improvement is $+0.033$ and the median $+0.031$, with the smallest margin
$+0.006$ and the largest $+0.061$.

\paragraph{Selection-strategy ablation, per domain.}
Table~\ref{tab:strategies} gives the per-domain breakdown for the hand-crafted
pool. Four of the nine domains are entirely insensitive to the strategy, and the
aggregate spread is driven mostly by facility (87 series) and sensor (39
series). The oracle column picks the best of the three per domain with hindsight
and is not an achievable configuration; it bounds what a better selection rule
could buy at $0.551$ per-series, only $0.008$
above the best achievable one.

\begin{table}[H]
\centering
\setlength{\tabcolsep}{3.5pt}
\small
\begin{tabular}{@{}lccccc@{}}
\toprule
Domain & $n$ & greedy & top-$k$ & mRMR & oracle \\
\midrule
environment    & 18 & 0.360 & 0.360 & 0.360 & 0.360 \\
facility       & 87 & 0.604 & 0.565 & 0.656 & 0.656 \\
finance        & 8  & 0.768 & 0.780 & 0.780 & 0.780 \\
human activity & 43 & 0.238 & 0.239 & 0.239 & 0.239 \\
medical        & 47 & 0.593 & 0.538 & 0.538 & 0.593 \\
sensor         & 39 & 0.686 & 0.758 & 0.754 & 0.758 \\
synthetic      & 39 & 0.483 & 0.488 & 0.488 & 0.488 \\
traffic        & 5  & 0.440 & 0.415 & 0.411 & 0.440 \\
web services   & 64 & 0.534 & 0.536 & 0.535 & 0.536 \\
\midrule
per-series     & 350 & 0.529 & 0.521 & 0.543 & 0.551 \\
per-domain     & 9   & 0.523 & 0.520 & 0.529 & 0.539 \\
\bottomrule
\end{tabular}
\caption{Selection-strategy ablation on the hand-crafted pool, per domain
(VUS-PR on the evaluation split).}
\label{tab:strategies}
\end{table}

\newpage
\section{Generation Prompt}
\label{app:prompt}

The instruction below is across all domains, generators, and
seeds; only the attached plots change. It is issued as a single user turn whose
content interleaves the text fragments and the ten rendered plots in the order
shown, preceded by the system message.

\begin{listing}[h]
\begin{lstlisting}[basicstyle=\scriptsize\ttfamily]
[system]
You are an expert time-series analyst and Python
programmer specialising in unsupervised anomaly
detection. You design discriminative statistical
features computed on a single sliding window of a
univariate series.

[user]
Below are example sliding windows from this domain's
time series, shown as line plots (x = sample index,
y = value). All plots share the same y-axis so levels
and magnitudes are directly comparable across windows.
Note: window lengths VARY - each window spans one
dominant period of its series - so your features must
be length-agnostic (do not assume a fixed len(x)).

NORMAL windows (no anomaly present):
N1: <plot>   ...   N5: <plot>

ANOMALOUS windows (contain a labelled anomaly):
A1: <plot>   ...   A5: <plot>

Propose EXACTLY 10 Python feature functions that would
help a robust (median/MAD) detector separate anomalous
windows from normal ones in THIS domain. Study the
plots and target the failure modes you see (e.g. level
shifts, spikes, variance changes, shape/periodicity
changes).

Requirements for each function:
  - signature: def feat_<snake_name>(x):  where x is a
    1-D numpy array (one window)
  - use numpy, available as `np` (do NOT write any
    import statements)
  - return a single finite Python float
  - be deterministic, pure, and robust to short windows
    (len(x) can be as low as 2)
  - no I/O, no randomness, no global state
  - give the 10 functions distinct, descriptive names

Return ONLY one ```python code block containing the 10
function definitions and nothing else (no prose, no
example calls).
\end{lstlisting}
\caption{The generation prompt in full. The domain is never named, and no
dataset identifier, detector description, score, or evaluation series is
included; the model reasons from the plots alone.}
\label{lst:prompt}
\end{listing}

\end{document}